\documentclass[10pt,journal,compsoc]{IEEEtran}

\ifCLASSOPTIONcompsoc
  \usepackage[nocompress]{cite}
\else
  \usepackage{cite}
\fi
\usepackage{siunitx}
\usepackage{gensymb}
\usepackage{makecell}

\usepackage{subfigure}
\usepackage{mathtools}
\usepackage{hyperref}
\usepackage[flushleft]{threeparttable}
\usepackage{multirow}
\usepackage{array}
\usepackage[dvipsnames]{xcolor}
\ifCLASSINFOpdf
\else
\fi

\usepackage{graphicx}

\begin{document}
%
\title{Synfeal: A Data-Driven Simulator for \\ End-to-End Camera Localization}
%
%
%
%

\author{Daniel~Coelho,
        Miguel~Oliveira,
        and~Paulo~Dias
\IEEEcompsocitemizethanks{\IEEEcompsocthanksitem Daniel Coelho is with the Department of Mechanical Engineering, University of Aveiro, 3810-193 Aveiro, Portugal, and with the Intelligent System Associate Laboratory (LASI), Institute of Electronics and Informatics Engineering of Aveiro (IEETA), University of Aveiro, 3810-193 Aveiro, Portugal.
\protect\\
E-mail: danielsilveiracoelho@ua.pt
\IEEEcompsocthanksitem Miguel Oliveira is with the Department of Mechanical Engineering, University of Aveiro, 3810-193 Aveiro, Portugal, and with the Intelligent System Associate Laboratory (LASI), Institute of Electronics and Informatics Engineering of Aveiro (IEETA), University of Aveiro, 3810-193 Aveiro, Portugal.
\protect\\
E-mail: mriem@ua.pt.
\IEEEcompsocthanksitem Paulo Dias is with the Department of Electronics, Telecommunications and Informatics, University of Aveiro, 3810-193 Aveiro, Portugal, and with the Intelligent System Associate Laboratory (LASI), Institute of Electronics and Informatics Engineering of Aveiro (IEETA), University of Aveiro, 3810-193 Aveiro, Portugal.
\protect\\
E-mail: paulo.dias@ua.pt.}
}

\IEEEtitleabstractindextext{%
\begin{abstract}
Collecting real-world data is often considered the bottleneck of Artificial Intelligence, stalling the research progress in several fields, one of which is camera localization. End-to-end camera localization methods are still outperformed by traditional methods, and we argue that the inconsistencies associated with the data collection techniques are restraining the potential of end-to-end methods.
Inspired by the recent data-centric paradigm, we propose a framework that synthesizes large localization datasets based on realistic 3D reconstructions of the real world. Our framework, termed \textbf{Synfeal}: \textit{Synthetic from Real}, is an open-source, data-driven simulator that synthesizes RGB images by moving a virtual camera through a realistic 3D textured mesh, while collecting the corresponding ground-truth camera poses. The results validate that the training of camera localization algorithms on datasets generated by \textbf{Synfeal} leads to better results when compared to datasets generated by state-of-the-art methods. Using \textbf{Synfeal}, we conducted the first analysis of the relationship between the size of the dataset and the performance of camera localization algorithms. Results show that the performance significantly increases with the dataset size. Our results also suggest that when a large localization dataset with high quality is available, training from scratch leads to better performances. \textbf{Synfeal} is publicly available at \url{https://github.com/DanielCoelho112/synfeal}.


\end{abstract}

}

\maketitle

\IEEEdisplaynontitleabstractindextext

%
\IEEEpeerreviewmaketitle

\section{Introduction} \label{lab:introduction}

\IEEEPARstart{C}{amera} localization refers to the problem of estimating a camera pose (position and orientation) from an image \cite{9229078}. It is of great importance to many systems, such as autonomous driving \cite{9411961}, virtual reality  devices \cite{8798257}, delivery drones \cite{9357892}, among others. The classic approach to this problem, known as 3D structure-based methods, establishes correspondences between 2D features in a query image and 3D points in a 
point cloud via descriptor matching \cite{9229078}. The point cloud is usually created using structure from motion (SfM). These 3D structure-based approaches are unable to find correspondences in all possible scenarios, in particular in scenarios with repetitive or textureless surfaces, occlusions, or large viewpoint changes \cite{Melekhov2017}. 

Contrary to 3D structure-based methods, end-to-end camera localization methods aim to train a neural network to regress the 6DoF camera pose from an RGB image in an end-to-end manner \cite{9347540}. 
A critical element in end-to-end camera localization, and often overlooked, is the collection of data. 
The ground truth, i.e, the camera pose associated with each image, is commonly acquired using SfM \cite{open_mvg,Lee2021} or SLAM \cite{Shotton2013,Henry2014,Lee2021} techniques. The accuracy and consistency of the ground truth is critical for any supervised system. Despite this, the literature reports innumerous situations where these techniques produce significant errors in the camera pose estimation: scenes with large volume, scenes with repetitive and textureless surfaces, or when the camera motion is large \cite{OLIENSIS2000172,BUENO2015137,8468281}. As such, one must consider the possibility that the errors associated with data collection introduce inconsistencies into the localization datasets, increasing the level of randomness in the function to be learned, which is then translated into worst localization performances. Moreover, current data collection techniques usually involve a human or a robot moving around the environment to collect the images, making it a cumbersome and time consuming to collect large datasets that capture a significant amount of the details of the the scene.


In line with other fields, the research paradigm in localization has focused on the refinement of models, neglecting the importance of the quality and dimension of the datasets.
Most of the research papers found in this field focus on developing robust algorithms to cope with the inconsistencies of the most common datasets, as is the case of the 7-Scenes dataset \cite{Shotton2013}. This paradigm has led to marginal performance increases over the years, which leads us to question whether this is the best approach.
Recently, Artificial Intelligence (AI) pioneer Andrew Ng argued that AI researchers should do precisely the opposite: fix the models, work on the data \cite{mdatacentric}. He advocates for data-centric AI, instead of model-centric AI, and claims that turning our attention to high-quality data that is consistently labeled will unlock the true potential of AI \cite{data-centric}.


Aligned with the data-centric paradigm, several AI researchers have turned their attention to synthetic data in order to overcome the problems related with real world datasets. For example, research on autonomous driving has sharply accelerated due to the usage of the self-driving car simulator CARLA \cite{carla}. The same has happened in the robotics field with MuJoCo \cite{mujoco}. However, the carry over from simulation to reality poses challenging obstacles due to the sim-to-real gap \cite{Muller2018}.

On the one hand, synthetic data generation is faster, free of inconsistencies, and more scalable than collecting real-world data. On the other, synthetic data often lacks the realism of real-world data, which compromises the real world performance of systems trained with simulated data.

With the goal of combining the best of both worlds, we propose a data-driven simulation framework that produces synthetic datasets from realistic 3D reconstructions of the scenes. Our framework, termed \textbf{Synfeal}: \textit{\underline{Syn}thetic \underline{f}rom R\underline{eal}}, synthesizes RGB images by moving a virtual camera through a realistic 3D textured mesh, while collecting the corresponding ground truth camera poses. Although this paper focuses on the application of \textbf{Synfeal} in the localization domain, the scope of \textbf{Synfeal} is much broader. The idea of producing synthetic data using 3D reconstructions of the real world can be employed in several fields.
In summary, the key contributions of this paper are as follows: 

\begin{itemize}
    \item[1)] \textbf{Synfeal}, an open-source, data-driven simulator that produces large localization datasets free of inconsistencies;
    
    \item[2)] Experimental validation that the training of camera localization algorithms on datasets with the ground truth provided by \textbf{Synfeal} leads to better performances than using datasets with the ground truth provided by state of the art methods;
            
    \item[3)] An analysis of the impact of the size of the datasets on the performance of end-to-end camera localization algorithms. Results show that large localization datasets without inconsistencies greatly improve the localization performance;
    
    \item[4)] Experimental validation that when a large localization dataset with high quality is available, training from scratch leads to better results in the localization domain.
\end{itemize}

\section{Related Work}

This section is divided into two parts: end-to-end camera localization and learning in simulation. 

\textbf{End-to-end camera localization:} PoseNet \cite{posenet} was the first work to show that Deep Learning (DL) can be used to directly regress the camera pose from a single RGB image, in a end-to-end manner. PoseNet uses GoogleNet \cite{googlenet}, pre-trained on large-scale image classification datasets, as the feature extractor, and then applies fully-connected layers to output the camera poses. The authors further improved this work with Bayesian neural networks to estimate the uncertainty of the predicted camera pose \cite{baysean}. Following works have incorporated Long Short-Term Memory units (LSTMs) into the pipeline. Walch et al., proposed PoseLSTM which employs LSTMs to reduce the feature dimensionality in a structured way \cite{poseLSTM}, and Clark et al. proposed VidLoc, in which the LSTMs are used to predict camera poses from videos \cite{vidloc}. Wu et al. proposed BranchNet that predicts the position and orientation by two different branches after the $5^{th}$ inception module \cite{branchnet}.
Melekhov et al., inspired by recent works on image restoration and semantic segmentation, proposed Hourglass, which is a symmetric encoder-decoder neural network to estimate the camera poses \cite{hourglass}. The approaches discussed above have combined the position and orientation objective functions into a single loss function, using a linear weighted sum. This has as a downside an additional hyperparameter to balance the position and orientation. To overcome this limitation, Kendall et al. proposed a novel loss function that is able to learn a weighting between the position and orientation objective functions \cite{posenet2}. The authors also introduced a loss function based on geometric reprojection error, however, in most cases the convergence is not attained. Finally, Wang et al. proposed AtLoc, which uses an attention module to force the network to focus on geometrically robust objects and features \cite{atloc}. All aforementioned approaches leverage transfer learning from large classification datasets, mainly due to the difficulty of acquiring large datasets in the real world \cite{poseLSTM}. Shavit et al. performed a comparative analysis of the performance of end-to-end camera localization methods in the 7-Scenes dataset, where results show that these methods are still outperformed by 3D structure-based methods by a large margin \cite{Shavit2019IntroductionTC}. For instance, considering the Fire scene, the average of the median error of the approaches previously mentioned was 0.37\unit{\meter} in position and 12º in orientation, which are far from the the 3D structure-based method (Active Search): 0.03\unit{\meter} and 1.5º \cite{Shavit2019IntroductionTC}. A possible cause for this considerable difference might the quality of the dataset itself and not the models. The ground truth poses were computed using KinectFusion \cite{kinectfusion}, which, as previously mentioned, can introduce inconsistencies into the dataset, increasing therefore difficulty of the learning process. An alternative to overcome this problem is to proceed with the collection of the dataset in a simulated environment, however, as will be explained below, there are some limitations that must be addressed.

\textbf{Learning in simulation:} Simulation has been the preferred solution by most researchers to overcome the limitations of collecting data in the real world. Traditional simulators are called model-based and they relied on artificially designed environments. In recent years, several model-based simulators have proved their value in different fields, as is the case of autonomous driving \cite{carla,9294422} and robotics \cite{mujoco, gazebo}. These simulators rely on video game rendering platforms, but still lack the realism necessary to directly transfer a learned policy to the real world \cite{vista2}. Recently, data-driven simulators were proposed to address the lack of realism of model-based simulators. Data-driven simulators use data from the real world to reconstruct virtual worlds of the scene before synthesizing novel views \cite{vista2}. Examples of data-driven simulators include \textbf{Gibson} \cite{gibson}, \textbf{Habitat 2.0} \cite{habitat, habitat2}, and \textbf{VISTA 2.0} \cite{vista, vista2}. \textbf{Gibson} enables the development of real world perception for active agents \cite{gibson}. \textbf{Gibson} is used to train tasks related with navigation, such as local planning and visual navigation. 
It uses a set of sparse RGB-D images and renders a panorama from an arbitrary novel viewpoint. According to the selected viewpoint, the local gaps are interpolated with bilinear sampling, and then a neural network filter is used to fix the artifacts.  One limitation of \textbf{Gibson} concerns the difference between the real images and the synthesized images, mainly due to geometric inconsistencies and texture seams. This limitation will certainly pose additional challenges to directly transfer a policy learned in simulation to the real world. \textbf{Habitat 2.0} is a platform for embodied AI research developed by Facebook. It enables the training of virtual agents in interactive 3D environments. \textbf{Habitat 2.0} is used to train tasks related with assistive robots, as is the case of picking/placing objects and opening/closing containers \cite{habitat2}. \textbf{Habitat 2.0} uses expert artists to model 3D environments to replicate the real environments. This approach to recreate the real environments has some limitations: a) a manually designed environment will always have some discrepancies when compared with the real environment, b) the need to manually design the virtual environments will prevent researchers to use this simulator due to lack of expertise in 3D modeling, and c) a realistic 3D modeling of a scene is both expensive and time consuming. \textbf{VISTA 2.0} is a data-driven simulator for autonomous driving that synthesizes multimodal sensor information at novel viewpoints around the local trajectory of a previously collected dataset \cite{vista2}. Novel viewpoints are synthesized leveraging geometric transformations based on the movement of the vehicle. The main limitation of \textbf{VISTA 2.0} is that novel viewpoints are spatially restricted by the movement of the vehicle in the dataset.

\textbf{Synfeal} belongs to the group of data-driven simulators, but unlike the aforementioned simulators, it leverages state of the art 3D reconstruction technologies using 3D LiDARs and cameras.
By using realistic 3D reconstructions, \textbf{Synfeal} ensures that the synthesized data reflects the real world data, and enables the synthesis of data in any viewpoint of the environment. 
Furthermore, \textbf{Synfeal} also distinguishes itself from the existing data-driven simulators by being the first applied to the localization field.

\section{Synthetic from Real}

\begin{figure}[!t]
\centering
\includegraphics[width=7.8cm]{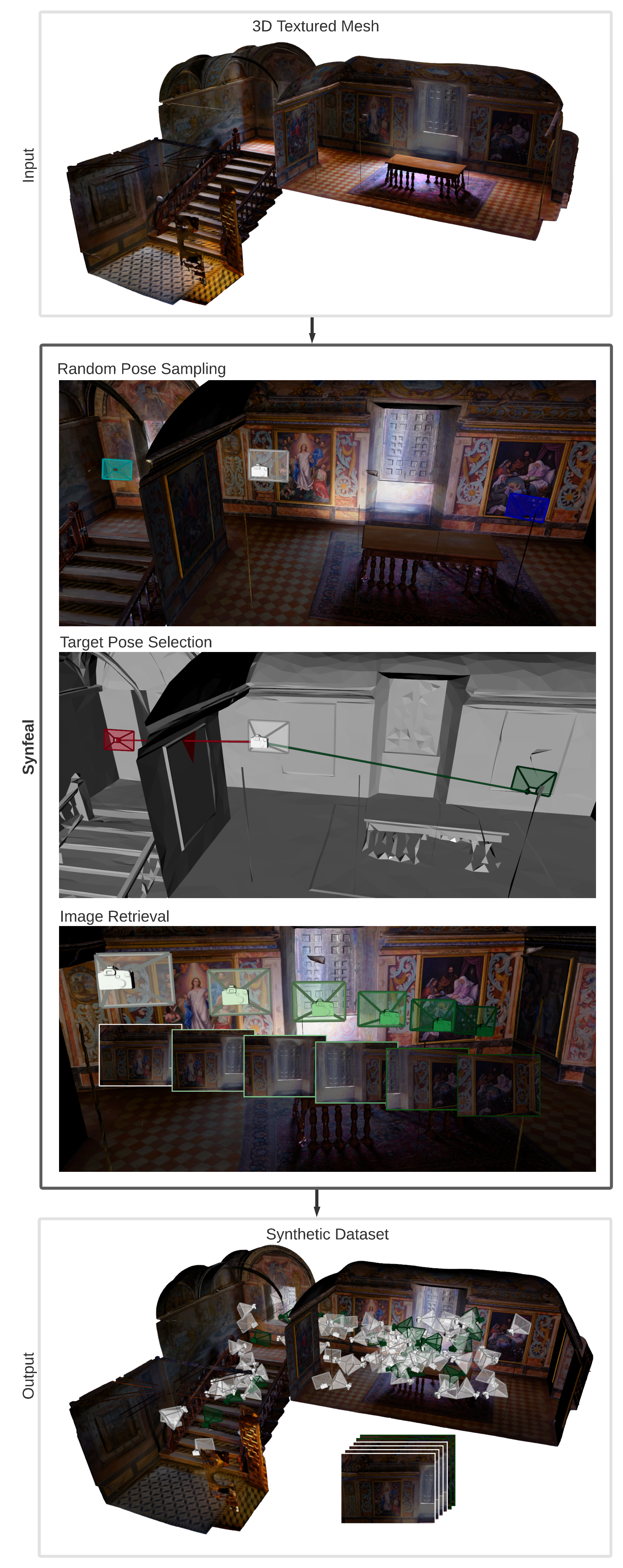}
\caption{Overview of \textbf{Synfeal}. The input is a 3D textured mesh that is used to generate a synthetic dataset for camera localization. \textbf{Synfeal} performs three sequential tasks: Random Pose Sampling, Target Pose Selection, and Image Retrieval. Random Pose Sampling generates target pose candidates (shown in cyan and blue), and then Target Pose Selection selects the target pose (highlighted in green). Finally, Image Retrieval generates a path from the initial pose (shown in white) to the target pose and collects RGB images and their corresponding poses along the way. Each time the three tasks are executed, the data corresponding to one path is collected. To collect a dataset that covers the entire scene, the tasks are repeated iteratively. A video of the data collection is available at \url{https://www.youtube.com/watch?v=sRxalb6BoFs&ab}.}
\label{fig:arch}
\end{figure}

\textbf{Synfeal} is a data-driven simulator that synthesizes RGB images by moving a virtual camera through a realistic 3D textured mesh. The architecture of the framework is depicted in Figure \ref{fig:arch}. \textbf{Synfeal} uses Open Dynamics Engine (ODE) as the physics engine, and Object-Oriented Graphics Rendering Engine (OGRE) as the 3D graphics engine bundled into the Gazebo simulator \cite{gazebo}. The source code of \textbf{Synfeal} is publicly available at \url{https://github.com/DanielCoelho112/synfeal}.

The core idea of \textbf{Synfeal} is to use realistic 3D reconstructions of the environment to produce synthetic datasets that can be used to train end-to-end camera localization algorithms, and then deploy them in the real world. In recent years, the research in 3D reconstruction has clearly increased, especially in 3D reconstruction using 3D LiDARs and cameras \cite{9311752,9189271,pami_3d}, leading to the creation of realistic 3D textured meshes. Replacing the real environment with the 3D textured mesh presents two main advantages in the data collection stage:
access to the accurate ground truth poses which removes the inconsistencies from the dataset, and automatic data collection, enabling the creation of large datasets.
The first step of \textbf{Synfeal} is to parameterize the virtual camera that will be responsible for collecting the RGB images. The virtual camera should have the same parameters as the real camera, for the synthetic images to be as similar as possible to the real images. This can be achieved through a classic intrinsic camera calibration procedure.
The output of \textbf{Synfeal} is a synthetic dataset, following the format of the 7-Scenes dataset \cite{Shotton2013}.

Once the 3D reconstruction is completed and the virtual camera is parameterized, the objective is to move the virtual camera through the 3D textured mesh and synthesize images. A naive approach would be to randomly generate poses inside the 3D textured mesh and synthesize images in those poses. However, this approach is far from optimal for three reasons: the spatial continuity between frames, characteristic of SLAM techniques, would be lost which could compromise the employment of algorithms that use more than one frame as input; the camera poses could be in positions that are impossible to replicate in the real world, for instance, the camera pose could be randomly generated inside a wall; and finally, there is no guarantee that the synthesized images have enough information that can be used to compute the localization (see Figure \ref{fig:camera_viewpoint}), for example, when the synthesized image is too close to a wall, that image could belong to several poses, introducing ambiguities into the dataset.

In light of the problems enumerated above, the data collection procedure in \textbf{Synfeal} is carried out by applying a linear movement from the initial pose of the virtual camera, defined as $p_i$, to a target pose, denoted as $p_t$, and collecting RGB images along the way. To accomplish that, 
\textbf{Synfeal} performs three sequential tasks: Random Pose Sampling (Section \ref{sec:random_pose_sampling}), Target Pose Selection (Section \ref{sec:target_pose_selection}), and Image Retrieval (Section \ref{sec:image_retrieval}). 
Each time the three tasks are completed, the RGB images of the path between $p_i$ and $p_t$ are collected along with their corresponding ground truth poses. To produce a dataset that covers the entire scenario, multiple paths are necessary and therefore the tasks are repeated iteratively. The following sections discuss each task individually.

\subsection{Random Pose Sampling} \label{sec:random_pose_sampling}




The objective of this task is to randomly generate $k$ poses inside the volume of the 3D textured mesh, denoted as $\mathcal{M}$. Each sampled pose will be a target pose candidate, represented as a vector of 6 elements, defined by the $x$,$y$,$z$ Cartesian coordinates and the roll, pitch, and yaw angles, denoted as $\theta$, $\phi$, and $\gamma$, respectively. Each element of the pose is extracted from a uniform distribution:

\begin{equation}
\begin{split}
    v_c \sim {U} \{v_{min}, v_{max}\}~, \hspace{1cm} 
    & \forall v \in \{x, y, z, \theta, \phi, \gamma \}~, \\
    & \forall c \in [1,k]~,
\end{split}
\end{equation}

\noindent where $x_{min}$, $x_{max}$, $y_{min}$, $y_{max}$, $z_{min}$, and $z_{max}$ are the minimum and maximum values along the $X$,$Y$, and $Z$ dimensions of the volume occupied by $\mathcal{M}$, respectively, and $\theta_{min}$, $\theta_{max}$, $\phi_{min}$, $\phi_{max}$, $\gamma_{min}$, and $\gamma_{max}$ are the minimum and maximum values for the admissible roll, pitch, and yaw angles, respectively. Finally. $x_c$,$y_c$,$z_c$,$\theta_c$,$\phi_c$, and $\gamma_c$ represent the elements of the \textit{c}-th generated pose ($p_c$). While the Cartesian boundaries are automatically extracted from $\mathcal{M}$, the angles boundaries are user-defined, allowing the user to control the diversity of the orientation of the camera poses.

At the end of this task, we have a set of target pose candidates, denoted as $\mathcal{X}$: 

\begin{equation}
    \mathcal{X} = \Bigg\{ \Big\{x_c ,y_c ,z_c ,\theta_c, \phi_c, \gamma_c\Big\}, \forall c \in [1,k]  \Bigg\}~.
\end{equation}

In Figure \ref{fig:arch}, the component \textit{Random Pose Sampling} illustrates this process when k=2. The white frustum represents the initial pose of the virtual camera, while the cyan and blue frustum represent the target pose candidates. Each target pose candidate defines a different path in which data is to be collected, and therefore a specific criterion must be used to select the target pose from the set of target pose candidates.

\subsection{Target Pose Selection} \label{sec:target_pose_selection}


Since the target pose candidates were defined under the only requirement of being inside the volume of $\mathcal{M}$, paths between the initial pose and candidate poses might lead to frames that would be impossible to obtain in the real world, e.g., when the virtual camera moves through a wall or frames that are impossible to extract the localization from the image, e.g., when the camera's viewpoint is too close to an obstacle. 

To discard the target pose candidates that lead to such frames, we decimate $\mathcal{M}$ and use the Ray casting algorithm \cite{4368179}. Starting with the collision problem, a clear example is the cyan pose from Figure \ref{fig:arch}. In component \textit{Target Pose Selection}, the path between the initial pose and the cyan pose leads to a collision (highlighted in red), and therefore the cyan should not be considered an admissible candidate. To accomplish that, for each target pose candidate, $p_c$, we compute the distance of the first collision of the ray that starts in $p_i$ and heads towards $p_c$, denoted as $R(p_i,p_c,\mathcal{M})$, and we discard $p_c$ from $\mathcal{X}$ if that distance is lower than the euclidean distance between the Cartesian coordinates of $p_i$ and $p_c$, implying that the path between $p_i$ and $p_c$ intercepts at least one triangle of $\mathcal{M}$.

The aforementioned condition removes all poses that involve collisions, however the problem related to the cases where the camera's viewpoint is too close to an object remains. Figure \ref{fig:camera_viewpoint} illustrates two cases where this problem occurs. Given the ambiguity of these images, they could be associated with more than one pose, which complicates the learning process. To overcome this problem, for each target pose candidate, we compute the distance of the first collision of the ray that starts in $p_c$, and has the orientation of the camera's viewpoint defined by $p_c$, denoted as $R(p_c,p_c,\mathcal{M})$, and then we discard $p_c$ if that distance is lower than a threshold denoted as $d$. If $d$ is sufficiently large, this condition ensures that the camera's viewpoint is far enough to capture relevant information. This filtering process can be formulated as:

\begin{figure}[t]

\begin{tabular}{cc}
\multicolumn{2}{c}{\includegraphics[width=8.4cm]{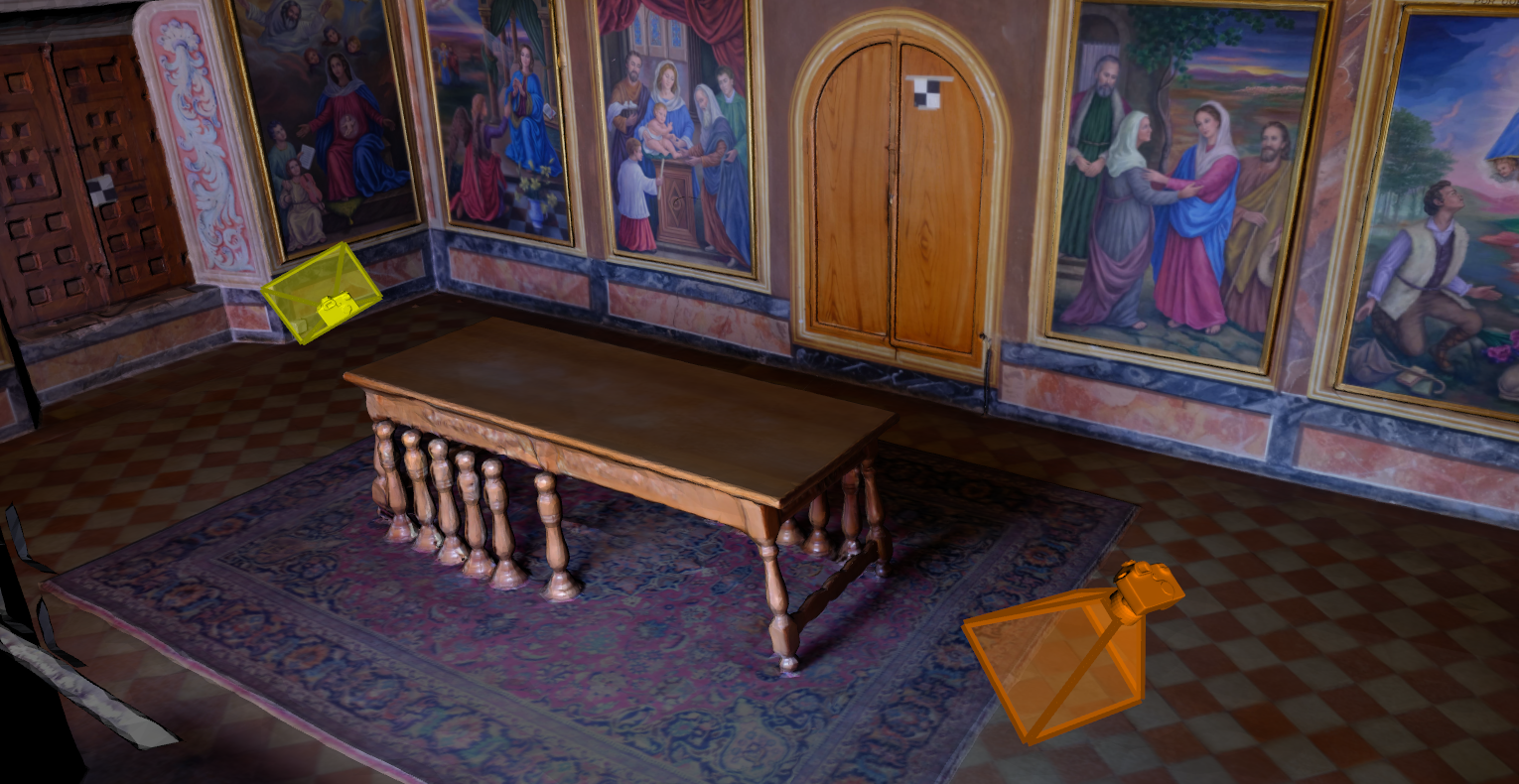}}\\
\multicolumn{2}{c}{({a})}\\
 \includegraphics[width=4cm]{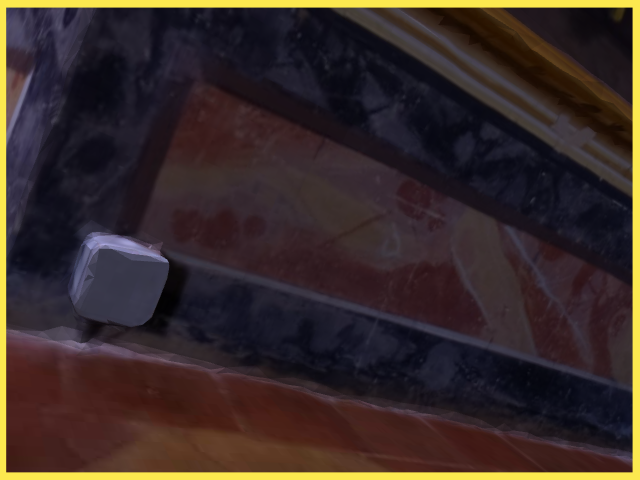}&
  \includegraphics[width=4cm]{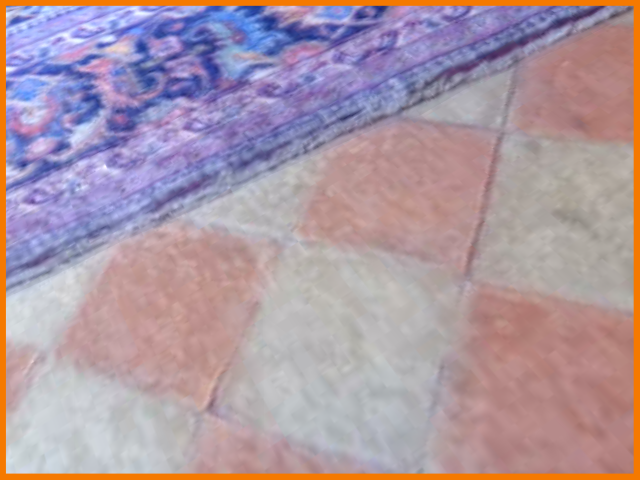}\\
({b})&({c})\\
\end{tabular}
\caption{Problem when the camera's viewpoint is too close to an obstacle. (a) represents the frustum of two poses where this problem occurs, and (b) and (c) depict the images generated by the yellow and orange frustum, respectively.}
\label{fig:camera_viewpoint}
\end{figure}


\begin{equation}
\begin{split}
    \mathcal{X}_a = 
    & \Bigg\{ p_c \in \mathcal{X} : \left(R(p_i, p_c, \mathcal{M}) > \Big\lVert p_i - p_c \Big\rVert \right)  
    \\ 
    & \wedge R(p_c,p_c,\mathcal{M}) > d   \Bigg\}~,
\end{split}
\end{equation}

\noindent where $\lVert . \rVert$ denotes the $L_2$ norm, and $\mathcal{X}_a$ corresponds to the subset of $\mathcal{X}$ that contains only the admissible target poses. To select the target pose from the admissible ones, the criterion used was to select the admissible pose that is farthest from the initial pose:

\begin{equation}
    p_t = \operatorname*{arg\,max}_{p_a} \Big( \lVert p_i - p_a \rVert \Big), \forall p_a \in \mathcal{X}_a~,
\end{equation}

\noindent where $p_t$ is the target pose. This criterion was employed in order to maximize the number of frames collected in each path.

\subsection{Image Retrieval} \label{sec:image_retrieval}


The objective of this task is to apply a linear movement to the virtual camera from $p_i$ to $p_t$ while collecting $n$ frames. In order to guarantee that different paths share the same distance between consecutive frames, $n$ must be defined as a function of the Euclidean distance between the Cartesian coordinates of $p_t$ and $p_0$ and of the distance between consecutive frames ($d$):

\begin{equation}
    n = \frac{\lVert p_t - p_i \rVert}{d}~,
\end{equation}

\noindent once $n$ is computed, we can determine the variation of each element of the poses ($\Delta v$) throughout the path:

\begin{equation}
    \Delta v = \frac{v_t - v_i}{n}~, \hspace{1cm} \forall v \in \{x, y, z, \theta, \phi, \gamma \}~,
\end{equation}

\noindent and finally, compute all elements of the pose for each frame:

\begin{equation}
\begin{split}
    v_{j} = v_{j-1} + \Delta v \cdot j~, \hspace{1cm} 
    & \forall v \in \{x, y, z, \theta, \phi, \gamma \}~, 
    \\
    & \forall j \in [1,n]~,
\end{split}
\end{equation}

\noindent where $v_j$ represents each element of the pose in all frames to collect.

The component \textit{Image Retrieval} in Figure \ref{fig:arch} shows the path from the initial pose (white frustum) to the target pose (dark green frustum) with n=6. The images synthesized in each frame are also shown. Once the target pose is reached, the random pose sampling is executed once more in order to repeat the cycle. The collection of data terminates when the predefined number of frames of the dataset is attained. A video of the collection of multiple paths, using k=1 and d = 2\unit{\centi\meter}, is available at \url{https://www.youtube.com/watch?v=sRxalb6BoFs&ab}.

\section{Results}

In this section, we analyze how the performance of end-to-end camera localization algorithms are influenced by: a) inconsistencies of the dataset, and b) the size of the dataset. Our goal is not to make a comparison between different models, i.e., we are not assessing which model performs best. Instead, we are focusing on what type of data creates an optimal learning environment for these models.

\begin{figure*}[t]
\centering
    \subfigure[]
    {\includegraphics[width=7cm]{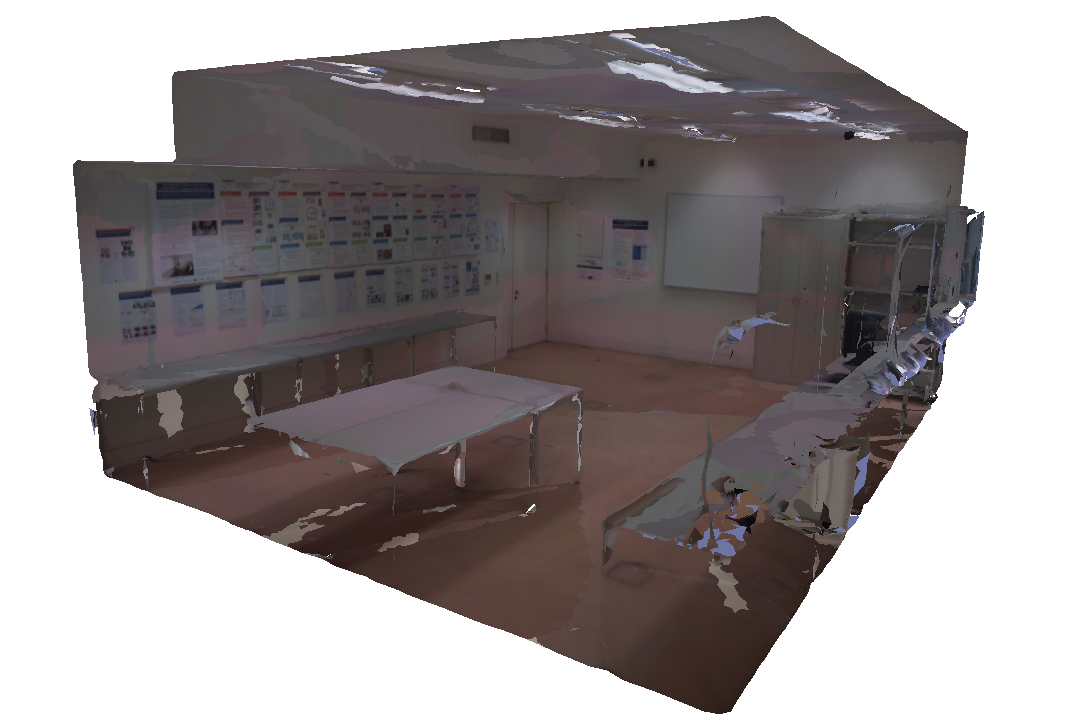}
    \label{fig:lab}
    }
    \subfigure[]
    {\includegraphics[width=9cm]{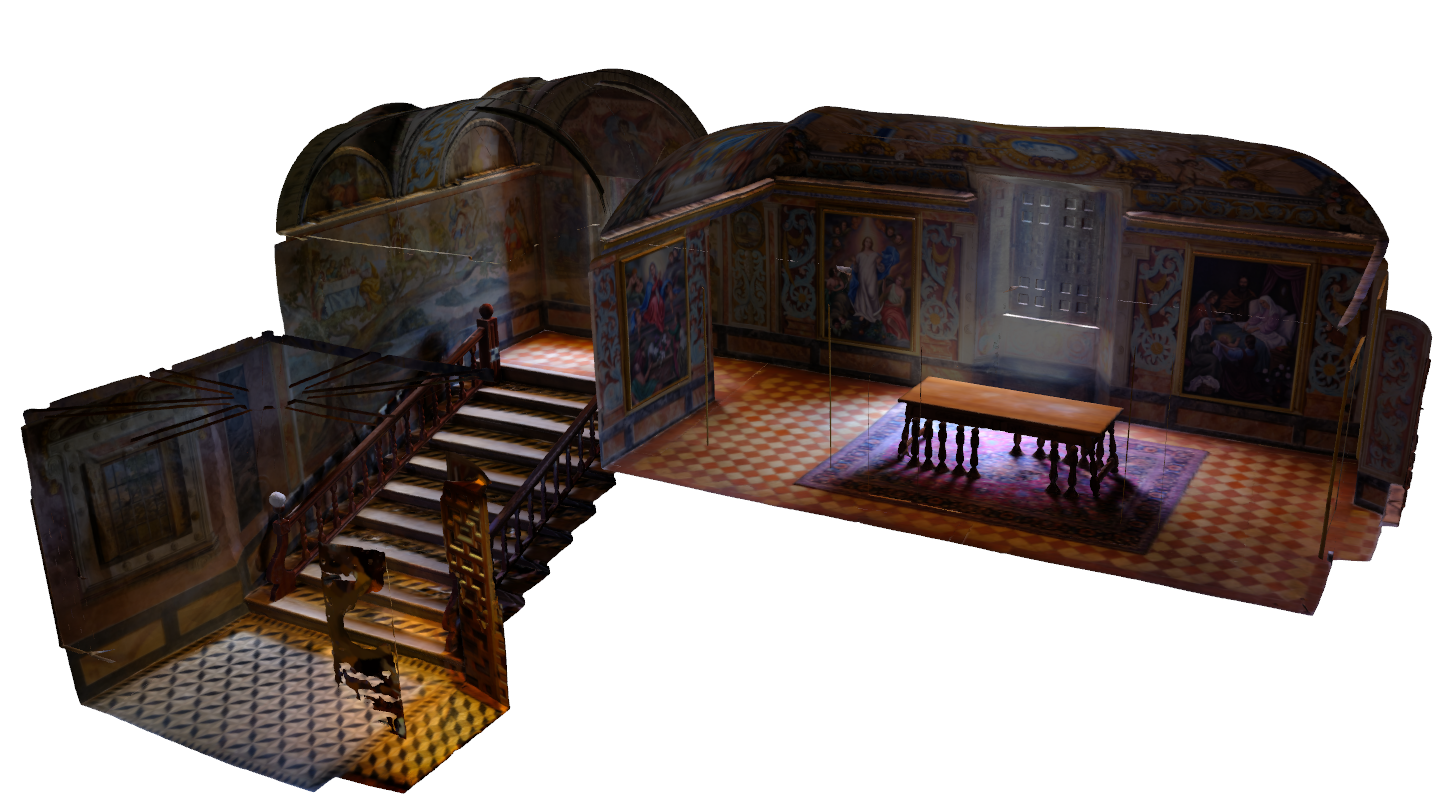}
    \label{fig:san}
    }
\caption{3D textured meshes used: (a) \textbf{Large Room} and (b) \textbf{Sanctuary}.}
\label{fig:confusion_matrices}
\end{figure*}

In order to increase the robustness of the analysis, we used two different 3D textured meshes representing challenging scenes: \textbf{Large Room} (Figure \ref{fig:lab}) and \textbf{Sanctuary} (Figure \ref{fig:san}). \textbf{Large Room} represents a laboratory of IEETA – Institute of Electronics and Informatics at University of Aveiro in Portugal, and the acquisition of the 3D textured mesh was carried out using the Leica BLK360 device. \textbf{Sanctuary} represents the Sanctuary of Las Virtudes, La Mancha in Spain, and the acquisition of the 3D textured mesh was carried out using a Faro Focus 3D-X330. The properties of both 3D textured meshes are described in Table \ref{tab:3d_textured_meshes}. In comparison with the 7-Scenes dataset, the scenes we are using are substantially more challenging. The average volume of the scenes in 7-Scenes is 6.37 $m^3$, which means that \textbf{Large Room} is 14 times larger, and \textbf{Sanctuary} is 29 times larger. Furthermore, our scenes, especially \textbf{Large Room}, contain repetitive and textureless structures that introduce ambiguity in the localization. All datasets created using these scenes are publicly available \href{https://uapt33090-my.sharepoint.com/:f:/g/personal/danielsilveiracoelho_ua_pt/EvcRerA_cptBk01uM8PwmH4BCIF97e6qRkm1mNc67gHY-Q}{here}.

In order to have a broad assessment of the influence of the quality of the data on the training of end-to-end camera localization algorithms, we implemented five state of the art end-to-end camera localization algorithms namely PoseNet \cite{posenet}, PoseLSTM \cite{poseLSTM}, Hourglass \cite{hourglass}, PoseNet2 \cite{posenet2}, and PoseNet2Res \cite{posenet2} (same architecture as PoseNet2, but with ResNet34 instead of GoogleNet). The algorithms were implemented using the details and hyperparameters  described in the original papers, and all preprocessing steps were replicated. We used the PyTorch DL library \cite{pytorch} and all models were trained on a NVIDIA RTX 2080 TI. With the exception of Section \ref{sec:pretrained}, all models were trained leveraging transfer learning from the ImageNet dataset \cite{5206848}.

The remainder of the section is organized as follows: Section \ref{sec:sota} compares the performance of the localization algorithms using a dataset with the ground truth provided by \textbf{Synfeal}, and using datasets with the ground truth provided by state of the art techniques such as SfM and SLAM. Section \ref{sec:inconsistencies} presents a detailed analysis on the influence of the inconsistencies of the dataset on the performance of the localization algorithms. Section \ref{sec:size} analyses the influence of size of the dataset, and finally, Section \ref{sec:pretrained} evaluates the impact of transfer learning on the performance of the algorithms.



\begin{table}[]
\centering
\caption{Properties of the 3D textured meshes used.}
\label{tab:3d_textured_meshes}
\begin{tabular}{ccc}
\hline
                                  & \textbf{Large Room} & \textbf{Sanctuary} \\ \hline
\# Vertices                       &     122 248         &   198 965           \\ 
\# Faces                          &     192 500         &   397 516           \\ 
Volume $[m^3]$                      &     89.4         &    183.18          \\ 
3D visualization                              &    \href{https://sketchfab.com/3d-models/room-024-2039d8fa7f64484681ab93f3a432dd31}{link}          &      \href{https://sketchfab.com/3d-models/sanctuary-of-las-virtudes-la-mancha-spain-810485ac7d564f0aac9052d5a9d513be}{link}        \\  \hline
\end{tabular}
\end{table}

\subsection{Comparison with state of the art data collection methods} \label{sec:sota}


This section aims to validate the main hypothesis in this paper: current data collection techniques in localization introduce inconsistencies into the datasets, which hampers the learning process. We compared \textbf{Synfeal} with three state of the art data collection methods in localization: SfM implemented using openMVG \cite{open_mvg} and two graph-based RGB-D SLAM approaches implemented in the RTAB-MAP system \cite{RTAB}: the first uses TORO \cite{toro} (RGB-D SLAM w$\!$/ TORO) as the graph optimization algorithm, whereas the second uses $g^{2}o$ \cite{g2o} (RGB-D SLAM w$\!$/ $g^{2}o$). 
Two different graph-based SLAM methods were used because graph-based SLAM is one of the most widely used SLAM methods \cite{graph-slam}.
We have also tried to use KinectFusion \cite{kinectfusion} but the results were not successful. As reported in \cite{Whelan2015RealtimeLD}, this method is not able to deal with repetitive structures and large scenes, which is the case of our scenes.

To ensure a fair comparison, we collected a synthetic dataset using \textbf{Synfeal} in \textbf{Large Room} with the size of (\textbf{tr}=6k, \textbf{te}=2.5k) - 6k for the train set and 2.5k for test set, and then we created four different datasets maintaining the same images but with different ground truth poses: a) the original dataset created by \textbf{Synfeal}, b) a dataset using SfM to provide the ground truth poses, c) a dataset using RGB-D SLAM w$\!$/ $g^{2}o$ to provide the ground truth poses, and d) a dataset using RGB-D SLAM w$\!$/ TORO to provide the ground truth poses. With these four datasets, we can use the end-to-end camera localization algorithms to assess which technique generates the dataset that best promotes the learning process. Note that to use the RGB-D SLAM, in addition to the RGB images, we also synthesized the depth images associated with the RGB images.

As mentioned above, \textbf{Large Room} contains several textureless and repetitive surfaces, which presented additional difficulties for the SfM and RGB-D SLAM approaches. None of these approaches were able to conduct a successful complete reconstruction of the scene, and therefore, were not able to produce the ground truth poses. In fact, this is consistent with the results reported in \cite{poseLSTM}, where the authors were also unable to use SfM or other techniques on challenging scenes to produce the ground truth poses. 
However, since the images used by SfM and the RGB-D SLAM methods were synthesized by \textbf{Synfeal}, we developed an automatic assistance mechanism to feed the accurate ground truth poses whenever those methods started to produce a considerable error.
Using SfM and the RGB-D SLAM methods, we processed the 8.5k images using batches of 250, and then we compared the camera pose estimation of the last image of the batch with the ground truth poses provided by \textbf{Synfeal}. If the position error was greater than 20\unit{\centi\meter}, or the orientation error greater than 10º, we replaced the last camera pose estimation with the one provided by \textbf{Synfeal}.
This mechanism prevents the accumulation of drifting errors, which is one of the reasons why these methods often fail.
Using SfM, 3 out of the 64 batches were assisted by \textbf{Synfeal}, while with RGB-D SLAM w$\!$/ TORO were 6 and with RGB-D SLAM w$\!$/ $g^{2}o$ were 4. In general, the number of batches assisted were low, however, they were critical due to the error propagation.
With this guidance process, we ensure that the ground truth poses obtained using SfM and RGBD-SLAM are reasonable and thus admissible for comparison with \textbf{Synfeal}. 

Table \ref{tab:comparison_sota} shows the performance of the aforementioned end-to-end camera localization methods for each dataset generated by the different data collection techniques. Following \cite{posenet,poseLSTM,hourglass,posenet2}, the performance numbers are reported in terms of the median translation and orientation errors. Without exception, the dataset produced by \textbf{Synfeal} leads to the best performances by a large margin of all algorithms, both in terms of position and orientation. For instance, when moving from the dataset provided by SfM to the dataset provided by \textbf{Synfeal}, PoseNet2 decreased the position and orientation error from 0.31\unit{\meter} and 16.57º to 0.14\unit{\meter} and 6.70º, which corresponds to a percentual variation of -55\% in position and -60\% in orientation - (\textbf{p}=-55\%, \textbf{o}=-60\%). All results using the datasets provided by both RGB-D SLAM techniques achieved similar performances, substantially worse than the ones provided by \textbf{Synfeal}. This table clearly validates our hypothesis: current data collection techniques in localization introduce inconsistencies into the datasets, which hampers the learning process.

\begin{table*}[t]
\centering
\begin{threeparttable}
\caption{Performance comparison of median localization error using five end-to-end camera localization algorithms on datasets produced by different data collection techniques. The scene used was \textbf{Large Room}. Best scores highlighted in bold.}
\label{tab:comparison_sota}
\begin{tabular}{llllll}
\hline
\begin{tabular}[l]{@{}l@{}}Ground truth\\ provided by:\end{tabular}            & PoseNet \cite{posenet}               & PoseLSTM \cite{poseLSTM}               & Hourglass \cite{hourglass}              & PoseNet2ResNet \cite{posenet2}         & PoseNet2 \cite{posenet2}               \\ \hline
SfM \cite{open_mvg}                & 0.34 m, 10.31º         & 0.31 m, 10.89º         & 0.34 m, 11.46º         & 0.33 m, 19.90º         & 0.31 m, 16.57º         \\ 
RGB-D SLAM w$\!$/ TORO \cite{RTAB} & 0.37 m, 10.15º         & 0.33 m, 9.75º          & 0.39 m, 14.33º         & 0.30 m, 13.18º         & 0.30 m, 10.90º         \\ 
RGB-D SLAM w$\!$/ g2o \cite{RTAB}  & 0.32 m, 9.75º          & 0.38 m, 11.46º         & 0.24 m, 11.58º         & 0.21 m, 15.48º         & 0.21 m, 12.60º         \\ 
Synfeal            & \textbf{0.29} m, \textbf{8.02}º & \textbf{0.24} m, \textbf{6.65}º & \textbf{0.20} m, \textbf{7.28}º & \textbf{0.17} m, \textbf{9.74}º & \textbf{0.14} m, \textbf{6.70}º \\ \hline
\end{tabular}
\end{threeparttable}
\end{table*}

Another interesting conclusion that can be extracted from Table \ref{tab:comparison_sota} concerns the impact that a data-centric approach has, in comparison to model-centric approaches. The camera localization algorithms are ordered chronologically from left to right, which means that if we analyse the percentual difference of the errors between the oldest algorithm (PoseNet) and the newest (PoseNet2), in the dataset provided by \textbf{Synfeal}, we get a decrease of (\textbf{p}=-51\%, \textbf{o}=-32\%). These values can be interpreted as the percentual improvement throughout the years using a model-centric approach. Conversely, using a data-centric approach - \textbf{Synfeal}, the percentual differences of the errors using PoseNet2, between SfM and \textbf{Synfeal} are (\textbf{p}=-55\%, \textbf{o}=-60\%), which are significantly larger. This suggests that, in this case, a data-centric approach has more impact than model-centric approaches.

\subsection{Influence of the inconsistencies of the dataset} \label{sec:inconsistencies}

In Section \ref{sec:sota}, results demonstrated that the common methods to acquire the ground truth poses in localization introduce inconsistencies into the datasets which undermines the performance of the localization algorithms. However, a more detailed analysis should be conducted to better understand the relationship between the inconsistencies, expressed in terms of median position and orientation errors, and the performance of algorithms. To accomplish this, we used the dataset generated by \textbf{Synfeal} in Section \ref{sec:sota} and created several datasets by introducing random variations of different levels of magnitude on the position and orientation of the ground truth poses. Table \ref{tab:inconsistencies} shows the impact that each dataset has on the performance of the algorithms. In general, when the inconsistencies are reflected more in the orientation, the performance of the algorithms decreases heavily on the orientation, and decreases slightly on the position. The opposite effect occurs when the inconsistencies are reflected more in the position. When the inconsistencies affect the dataset both in position and orientation the negative effects are combined which leads to a large decrease in performance, both in position and orientation. For example, with PoseNet2, a dataset with median position and orientation errors of 0.30\unit{\meter} and 20º, respectively, leads to an error increase of (\textbf{p}=65\%, \textbf{o}=79\%).

\begin{table*}[t]
\centering
\begin{threeparttable}
\caption{Performance comparison of median localization error using five end-to-end localization algorithms on datasets with different levels of \textbf{inconsistencies}. The scene used was \textbf{Large Room}. Best scores highlighted in bold.}
\label{tab:inconsistencies}
\begin{tabular}{rllllll}
\hline
\multicolumn{1}{l}{}                                                        & \multicolumn{1}{l}{Inconsistencies} & \multicolumn{1}{c}{PoseNet \cite{posenet}} & \multicolumn{1}{c}{PoseLSTM \cite{poseLSTM}} & \multicolumn{1}{c}{Hourglass \cite{hourglass}} & \multicolumn{1}{c}{Posenet2ResNet \cite{posenet2}} & PoseNet2 \cite{posenet2}               \\ \hline
\multirow{4}{*}{\begin{tabular}[c]{@{}r@{}}\rotatebox[origin=c]{90}{Ori.}  \end{tabular}} & \textbf{0} m, \textbf{0}º                    & \textbf{0.29} m, \textbf{8.02}º               & \textbf{0.24} m, \textbf{6.65}º                &  \textbf{0.20} m,  \textbf{7.28}º                 & \textbf{0.17} m, \textbf{9.74}º                      & \textbf{0.14} m, \textbf{6.70}º \\
                                                                            & 0 m, 5º                             & 0,30 m, 9.92º               & 0.36 m, 12.15º               & 0.24 m, 11.69º                & 0.18 m, 13.90º                     & 0,15 m, 10.32º         \\
                                                                            & 0 m, 10º                            & 0.31 m, 13.76º              & 0.33 m, 16.05º               & 0.24 m, 15.76º                & 0.18 m, 18.91º                     & 0.16 m, 15.48º         \\
                                                                            & 0 m,20º                             & 0.39 m, 33.82º              & 0.44 m, 38.40º               & 0.35 m, 38.41º                & 0.20 m, 33.80º                     & 0.17 m, 27.90º         \\ \hline
\multirow{4}{*}{\begin{tabular}[c]{@{}r@{}}\rotatebox[origin=c]{90}{Pos.}  \end{tabular}}    & \textbf{0} m, \textbf{0}º                    & \textbf{0.29} m, \textbf{8.02}º               & \textbf{0.24} m, \textbf{6.65}º                & \textbf{0.20} m, \textbf{7.28}º                 & \textbf{0.17} m, \textbf{9.74}º                      & \textbf{0.14} m, \textbf{6.70}º \\
                                                                            & 0.075 m, 0º                         & 0.30 m, 8.03º               & 0.26m, 6.70º                 & 0.24 m, 9.17º                 & 0.18 m, 9.75º                      & 0.18 m, 7.45º          \\ 
                                                                            & 0.15 m, 0º                          & 0.32 m, 8.03º               & 0.34 m, 9.28º                & 0.29 m, 10.78º                & 0.25 m, 9.92º                      & 0.22 m, 7.45º            \\
                                                                            & 0.30 m, 0º                          & 0.42 m, 8.04º               & 0.44 m, 8.60º                & 0.40 m, 8.20º                 & 0.39 m, 10.43º                     & 0.39 m, 8.60º          \\ \hline
\multirow{4}{*}{\rotatebox[origin=c]{90}{Pos.+Ori.}}                                                 & \textbf{0} m, \textbf{0}º                    & \textbf{0.29} m, \textbf{8.02}º               &    \textbf{0.24} m, \textbf{6.65}º                & \textbf{0.20} m, \textbf{7.28}º                 & \textbf{0.17} m, \textbf{9.74}º                      & \textbf{0.14} m, \textbf{6.70}º \\
                                                                            & 0.10 m, 0º                          & 0.35 m, 11.47º              & 0.38 m, 13.18º               & 0.24 m, 12.04º                & 0.21 m, 15.92º                     & 0.18 m, 0.75º          \\
                                                                            & 0.20 m, 10º                         & 0.45 m, 17.78º              & 0.51 m, 20.06º               & 0.36 m, 23.79º                & 0.30 m, 19.6º                      & 0.28 m, 17.77º         \\
                                                                            & 0.30 m, 20º                         & 0.50 m, 38.98º              & 0.52 m, 29.81º               & 0.42 m, 37.84                 & 0.40 m, 38.94º                     & 0.39 m, 31.5º        \\ \hline  
\end{tabular}
\end{threeparttable}
\end{table*}

\subsection{Influence of the size of the dataset} \label{sec:size}

Due to the intricate nature of collecting real world data, localization datasets are often small. For instance, the 7-Scenes dataset has, on average, 3.7 k frames for training and 2.4k frames for testing.
Conversely, \textbf{Synfeal} can produce large datasets in a relatively short amount of time. Using an Intel(R) Core(TM) i7-1065G7 8-Core processor and a NVIDIA GeForce MX350, \textbf{Synfeal} can collect data at a frequency of 5 frames per second. Combining this with the fact that \textbf{Synfeal} collects data automatically, \textbf{Synfeal} can fix the problem of small datasets in the localization field.

\begin{figure*}[t]
\centering
    \subfigure[]
    {\includegraphics[width=8.0cm]{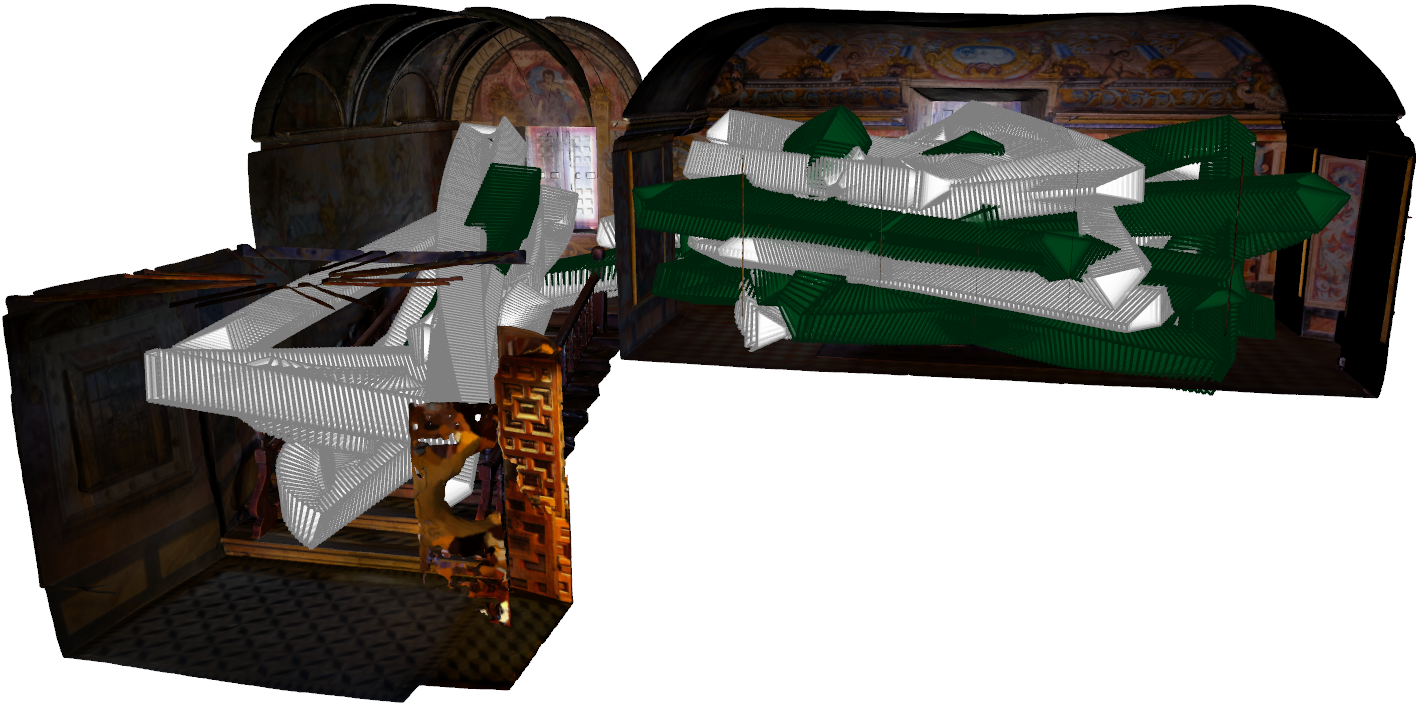}
    \label{fig:6k}
    }
    \subfigure[]
    {\includegraphics[width=8.0cm]{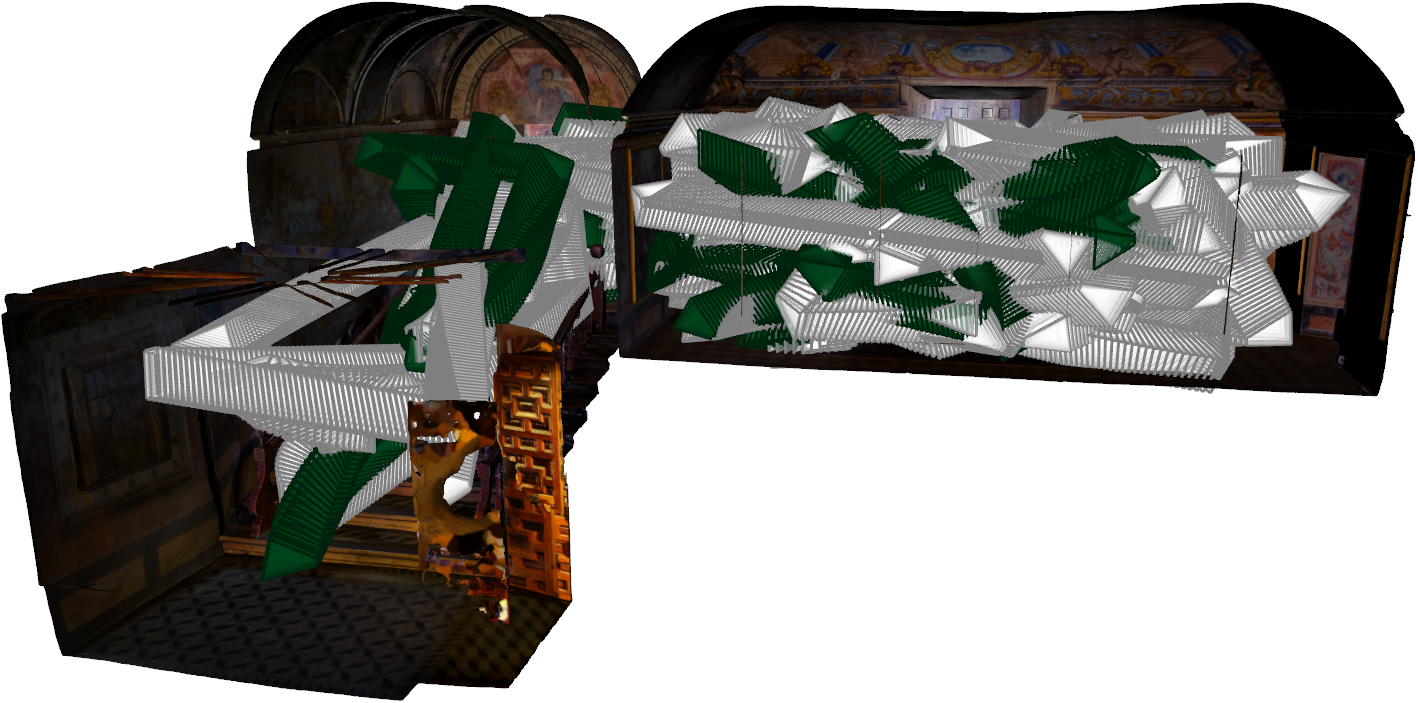}
    \label{fig:20k}
    }
    
    \subfigure[]
    {\includegraphics[width=8.0cm]{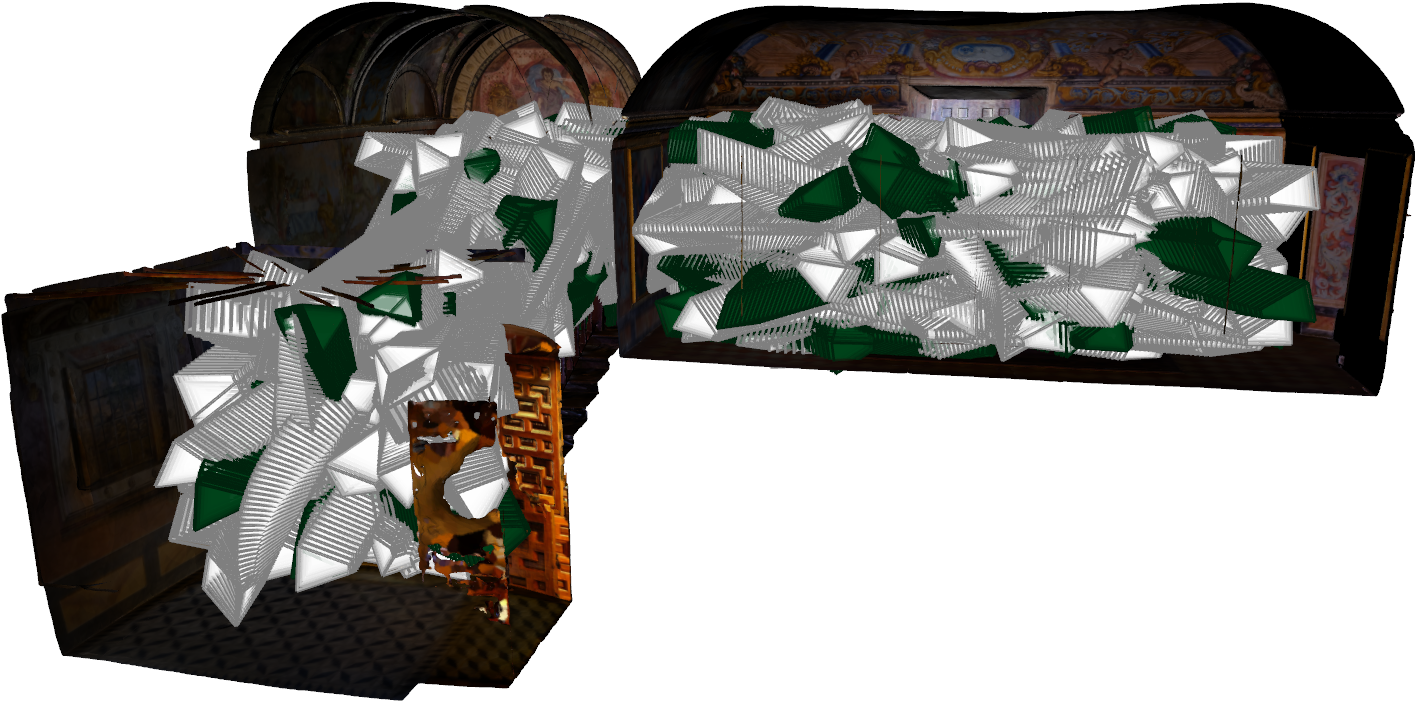}
    \label{fig:40l}
    }
    \subfigure[]
    {\includegraphics[width=8.0cm]{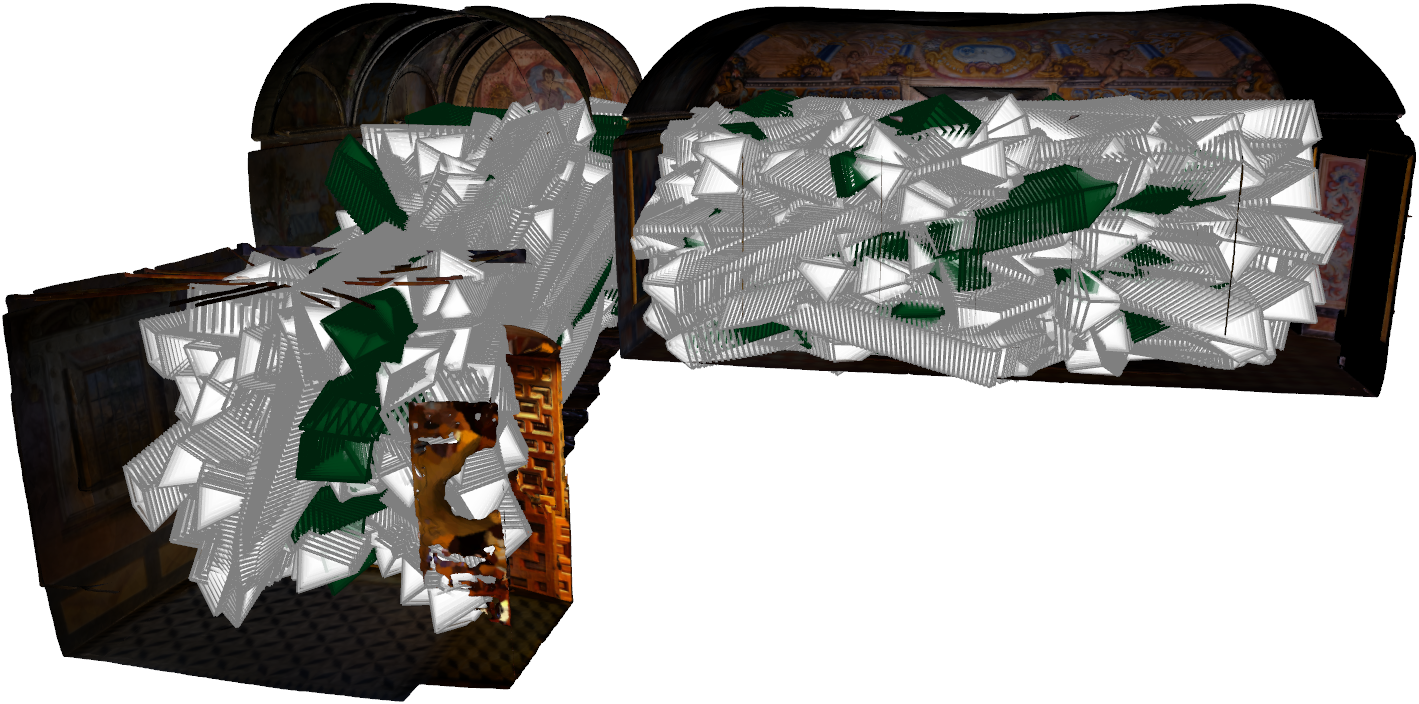}
    \label{fig:80k}
    }
\caption{Visual representation of datasets using \textbf{Sanctuary} with: (a) 6k for train and 2.5k for test, (b) 20k for train and 5k for test; (c) 40k for train and 10k for test, and (d) 95k for train and 20k for test. The camera poses identified with white frustum correspond to the train set, while the camera poses identified with dark green frustum correspond to the test set.}
\label{fig:size_of_san}
\end{figure*}

The scarcity of large localization datasets has prevented researchers from analyzing the relationship between the number of frames used and the performance of the localization algorithms. This section aims to address this issue. We created datasets with varied sizes for both \textbf{Large Room} and \textbf{Sanctuary}. As an example, Figure \ref{fig:size_of_san} depicts the camera poses of datasets with different size using \textbf{Sanctuary}. As the size of the dataset grows, more details are incorporated into the dataset, and therefore, it is expected that the algorithms achieve better performance.

Table \ref{tab:size} shows the performance of the end-to-end camera localization algorithms for each dataset using \textbf{Large Room} and \textbf{Sanctuary}. Considering the \textbf{Large Room} scene, the trend is clear: with the increase of the size of the dataset, the performance of the localization algorithms also improve. Moving from (\textbf{tr}=6k, \textbf{te}=2.5k), to (\textbf{tr}=80k, \textbf{te}=20k), results in a performance increase of, on average, (\textbf{p}=50\%, \textbf{o}=35\%). Considering PoseNet2, we were able to achieve localization errors of 0.04 m and 4.59º in a scene with 89.4 $m^3$, which is one order of magnitude below the state of the art, even using a far more challenging scene. 
This comparison clearly demonstrates the value that a data-centric approach, such as \textbf{Synfeal}, can offer to the field of localization.

The impact of the size of the dataset is even more noticeable in the \textbf{Sanctuary} scene, possibly due to the larger volume. As previously mentioned, common localization datasets have a size around the size of the first row of Table \ref{tab:size} - (\textbf{tr}=6k, \textbf{te}=2.5k). For large scenes, such as \textbf{Sanctuary},  Table \ref{tab:size} shows that this size is clearly insufficient, since the average performance attained by the localization algorithms is 0.70\unit{\meter} and 34.83º.
Moving from this small dataset to (\textbf{tr}=95k, \textbf{te}=20k), results in a performance increase of, on average, (\textbf{p}=72\%, \textbf{o}=75\%). These high percentages clearly demonstrate the usefulness of \textbf{Synfeal}, especially when dealing with large scenes.

While Sections \ref{sec:sota} and \ref{sec:inconsistencies} have demonstrated the benefits of using \textbf{Synfeal} from the perspective of eliminating inconsistencies, this section shows how \textbf{Synfeal} can significantly increase the performance of end-to-end camera localization algorithms by increasing the size of the dataset.

\begin{table*}[t]
\centering
\begin{threeparttable}
\caption{Performance comparison of median localization error using five end-to-end camera localization algorithms on datasets with \textbf{different size}. The top part of the table concerns the \textbf{Large Room} scene, and the bottom part concerns the \textbf{Sanctuary} scene. Best scores are highlighted in bold for both scenes.}
\label{tab:size}
\begin{tabular}{rllllll}
\hline
\multicolumn{1}{l}{}      & (\# train, \# test) & PoseNet \cite{posenet}              & PoseLSTM     \cite{poseLSTM}          & Hourglass \cite{hourglass}              & PoseNet2ResNet \cite{posenet2}          & PoseNet2    \cite{posenet2}           \\ \hline
\multirow{4}{*}{\rotatebox[origin=c]{90}{\textbf{\makecell{Large \\ Room}}}}    & (6k, 2.5k)            & 0.29 m, 8.02º          & 0.24 m, 6.65º          & 0.20 m, 7.28º          & 0.17 m, 9.74º           & 0.14 m, 6.70º          \\
                          & (20k, 5k)             & 0.25 m, 7.05º          & 0.21 m, 6.59º          & 0.12 m, 5.39º          & 0.10 m, 8.54º           & 0.09 m, 6.44º          \\
                          & (40k, 10k)            & 0.25 m, 5.73º          & 0.20 m, 6.19º          & 0.10 m, 4.58º          & 0.09 m, \textbf{7.22}º           & 0.07 m, 6.40º          \\
                          & (80k,20k)             & \textbf{0.20} m, \textbf{4.01}º & \textbf{0.15} m, \textbf{4.47}º & \textbf{0.09} m, \textbf{4.30}º & \textbf{0.08} m, 8.02º  & \textbf{0.04} m, \textbf{4.59}º \\ \hline
\multirow{4}{*}{\rotatebox[origin=c]{90}{\textbf{Sanctuary}}} & (6k, 2.5k)            & 0.80 m, 27.51          & 0.78 m, 39.40          & 0,71 m, 38.98º         & 0.57 m, 38.98º          & 0,62 m, 29.24º         \\
                          & (20k, 5k)             & 0.52 m, 15.93º         & 0.44 m, 12.38º         & 0.33 m, 15.36º         & 0.24 m, 15.36º          & 0.26 m, 15.48º         \\
                          & (40k, 10k)            & \textbf{0.32} m, 8.20º & 0.32 m, 10.00º         & 0.23 m, 7.97º          & 0.21 m, 17.66º          & 0.20 m, 12.60º         \\
                          & (95k, 20k)            & \textbf{0.32} m, \textbf{7.16}º & \textbf{0.24} m, \textbf{7.33}º & \textbf{0.17} m, \textbf{6.13}º & \textbf{0.15} m, \textbf{14.79}º & \textbf{0.12} m, \textbf{7.91}º \\ \hline
\end{tabular}
\end{threeparttable}
\end{table*}

\subsection{Assessment of the use of transfer learning} \label{sec:pretrained}

Training convolutional neural networks from scratch for the camera localization task has been considered impractical due to the lack of training data \cite{poseLSTM, hourglass}. Although transfer learning is considered a cornerstone of computer vision, recent works have indicated that in cases where a large-size dataset of high quality is available, training from scratch may produce better results \cite{transfer_learning,Shavit2019IntroductionTC}. Furthermore, the possibility of training from scratch can contribute to the progress in the field of localization, as it allows researchers to experiment with different architectures and sensor modalities.
This section aims to verify if large datasets with high quality can also benefit from training from scratch in the field of localization.

Table \ref{tab:tf} shows the performance of end-to-end camera localization algorithms without using transfer learning on datasets of different size using the \textbf{Large Room} scene. To accurately access the influence of transfer learning, Table \ref{tab:tf} should be compared with the top part of Table \ref{tab:size}, since the only difference is the usage of transfer learning. When considering the smallest dataset, i.e., (\textbf{tr}=6k, \textbf{te}=2.5k), the usage of transfer learning is highly recommended because all of the five algorithms suffered a significant performance decrease without the use of transfer learning. The same remains true when considering the dataset with (\textbf{tr}=20k, \textbf{te}=5k), however, the performance decrease is less noticeable. The turning point is the dataset with (\textbf{tr}=40k, \textbf{te}=10k), where the algorithms, with and without transfer learning, start to achieve the same level of performance. Finally, when considering the dataset with (\textbf{tr}=80k, \textbf{te}=20k), 3 out of the 5 camera localization algorithms benefit from training from scratch, which demonstrates that with large datasets, training from scratch leads to better results. Considering all algorithms, with and without transfer learning, PoseNet2 without transfer learning was the one that achieved the best results in the \textbf{Large Room}, with 0.03\unit{\meter} in position error and 4.50º in orientation error. These results clearly demonstrate that training from scratch, when a large dataset with high quality is available, leads to advantageous results. It is important to note that training from scratch implies more training time. For example, in the dataset with (\textbf{tr}=80k, \textbf{te}=20k), on average, the models with transfer learning were training during 72 hours whereas the models without transfer learning were training during 114 hours (around 58\% time increase).

\begin{table*}[t]
\centering
\begin{threeparttable}
\caption{Performance comparison of median localization error using five end-to-end camera localization algorithms, \textbf{without applying transfer learning}, on datasets with different size. The scene used was \textbf{Large Room}. The percentual variation considering the values when transfer learning was used (top part of Table \ref{tab:size}) is also represented. Red color indicates cases when the error using transfer learning is smaller, and green color indicates the opposite cases.}
\label{tab:tf}
\begin{tabular}{llllll}
\hline
\begin{tabular}[l]{@{}l@{}}\textbf{Large Room}\\ (\# train, \# test)\end{tabular} & PoseNet* \cite{posenet}                               & PoseLSTM* \cite{poseLSTM}                              & Hourglass* \cite{hourglass}                             & PoseNet2ResNet* \cite{posenet2}                       & PoseNet2*  \cite{posenet2}                            \\ \hline
                                                                  & 0.42 m, 16.05º                         & 0.39 m, 10.55º                         & 0.33 m, 12.90º                         & 0.24 m, 13.70º                        & 0.18 m, 8.94º                         \\
\multirow{-2}{*}{(6k,2.5k)}                                         & \multicolumn{1}{l}{\textcolor{red}{+44.8\%},\textcolor{red}{+100.1\%}} & \multicolumn{1}{l}{\textcolor{red}{+62.5\%}, \textcolor{red}{+58.6\%}}   & \multicolumn{1}{l}{\textcolor{red}{+65.0\%}, \textcolor{red}{+77.2\%}} & \multicolumn{1}{l}{\textcolor{red}{+41.2\%}, \textcolor{red}{+40.7\%}}  & \multicolumn{1}{l}{\textcolor{red}{+28.6\%},\textcolor{red}{+33.4\%}} \\ \hline
                                                                  & 0.31 m, 7.34º  & 0.28 m, 9.98º                          & 0.31 m, 17.02º                         & 0.12 m, 13.50º                        & 0.10 m, 7.85º                         \\
\multirow{-2}{*}{(20k, 5k)}                                         & \multicolumn{1}{l}{\textcolor{red}{+25.5\%},\textcolor{red}{+4.1\%}}    & \multicolumn{1}{l}{\textcolor{red}{+33.3\%}, \textcolor{red}{+51.4\%}}   & \multicolumn{1}{l}{\textcolor{red}{+158.3\%}, \textcolor{red}{+215.8\%}} & \multicolumn{1}{l}{\textcolor{red}{+20.0\%}, \textcolor{red}{+58.1\%}} & \multicolumn{1}{l}{\textcolor{red}{+16.3\%}, \textcolor{red}{+21.9\%}} \\ \hline
                                                                  & 0.30 m, 6.07º                          & 0.18 m, 5.15º                          & 0.15 m, 8.02º                          & 0.08 m, 10.32º                        & 0.07 m, 7.45º                         \\
\multirow{-2}{*}{(40k, 10k)}                                        & \multicolumn{1}{l}{\textcolor{red}{+21.5\%}, \textcolor{red}{+5.9\%}}   & \multicolumn{1}{l}{\textcolor{Green}{-10.0\%}, \textcolor{Green}{-16.8\%}} & \multicolumn{1}{l}{\textcolor{red}{+50.0\%}, \textcolor{red}{+75.1\%}}  & \multicolumn{1}{l}{\textcolor{Green}{-11.1\%}, \textcolor{red}{+42.9\%}} & \multicolumn{1}{l}{\textcolor{Green}{-5.4\%}, \textcolor{red}{+16.4\%}} \\ \hline
                                                                  & \textbf{0.29} m, \textbf{4.61}º                          & \textbf{0.14} m, \textbf{4.43}º                          & \textbf{0.09} m, \textbf{5.73}º                          & \textbf{0.05} m, \textbf{7.44}º                         &  \textbf{0.03} m,  \textbf{4.50}º                         \\
\multirow{-2}{*}{(80k, 20k)}                                        & \multicolumn{1}{l}{\textcolor{red}{+45.0\%}, \textcolor{red}{+15.0\%}}  & \multicolumn{1}{l}{\textcolor{Green}{-6.7\%}, \textcolor{Green}{-0.9\%}}   & \multicolumn{1}{l}{\textcolor{Green}{0.0 \%}, \textcolor{red}{+33.3 \%}}  & \multicolumn{1}{l}{\textcolor{Green}{-37.5\%}, \textcolor{Green}{-7.2\%}} & \multicolumn{1}{l}{\textcolor{Green}{-25\%}, \textcolor{Green}{-1.96\%}}  \\ \hline
\end{tabular}
\begin{tablenotes}
\footnotesize
\item * Without transfer learning.
\end{tablenotes}
\end{threeparttable}
\end{table*}

\section{Conclusion}

This paper proposed \textbf{Synfeal}, an open-source data-driven simulator that produces synthetic localization datasets from realistic 3D textured meshes. Our experiments validate that end-to-end camera localization algorithms can learn better using datasets with the ground truth provided by \textbf{Synfeal}, than using datasets with the ground-truth provided by state of the art methods, such as SfM and SLAM. Given the ability of \textbf{Synfeal} to produce large datasets in a short period of time, we also conducted the first analysis of the impact of the size of the dataset in the localization domain. Results demonstrated that with the increase of the size of the dataset, the performance of the algorithms increases substantially. Finally, our experiments also demonstrated that when a large dataset with high quality is available, training from scratch leads to better performances in the localization domain. In the future, we intend to extend the framework to other applications.

\ifCLASSOPTIONcompsoc
  \section*{Acknowledgments}
\else
  \section*{Acknowledgment}
\fi

This research was developed in the scope of the Augmented Humanity project [POCI-01-0247-FEDER-046103 and LISBOA-01-0247-FEDER-046103], financed by ERDF through POCI. It was also supported by FCT - Foundation for Science and Technology, in the context of Ph.D. scholarship 2022.10977.BD and by National Funds through the FCT - Foundation for Science and Technology, in the context of the project UIDB/00127/2020.

\ifCLASSOPTIONcaptionsoff
  \newpage
\fi

\bibliographystyle{IEEEtran}
\bibliography{references}

%

\begin{IEEEbiography}[{\includegraphics[width=1in,height=1.25in,clip,keepaspectratio]{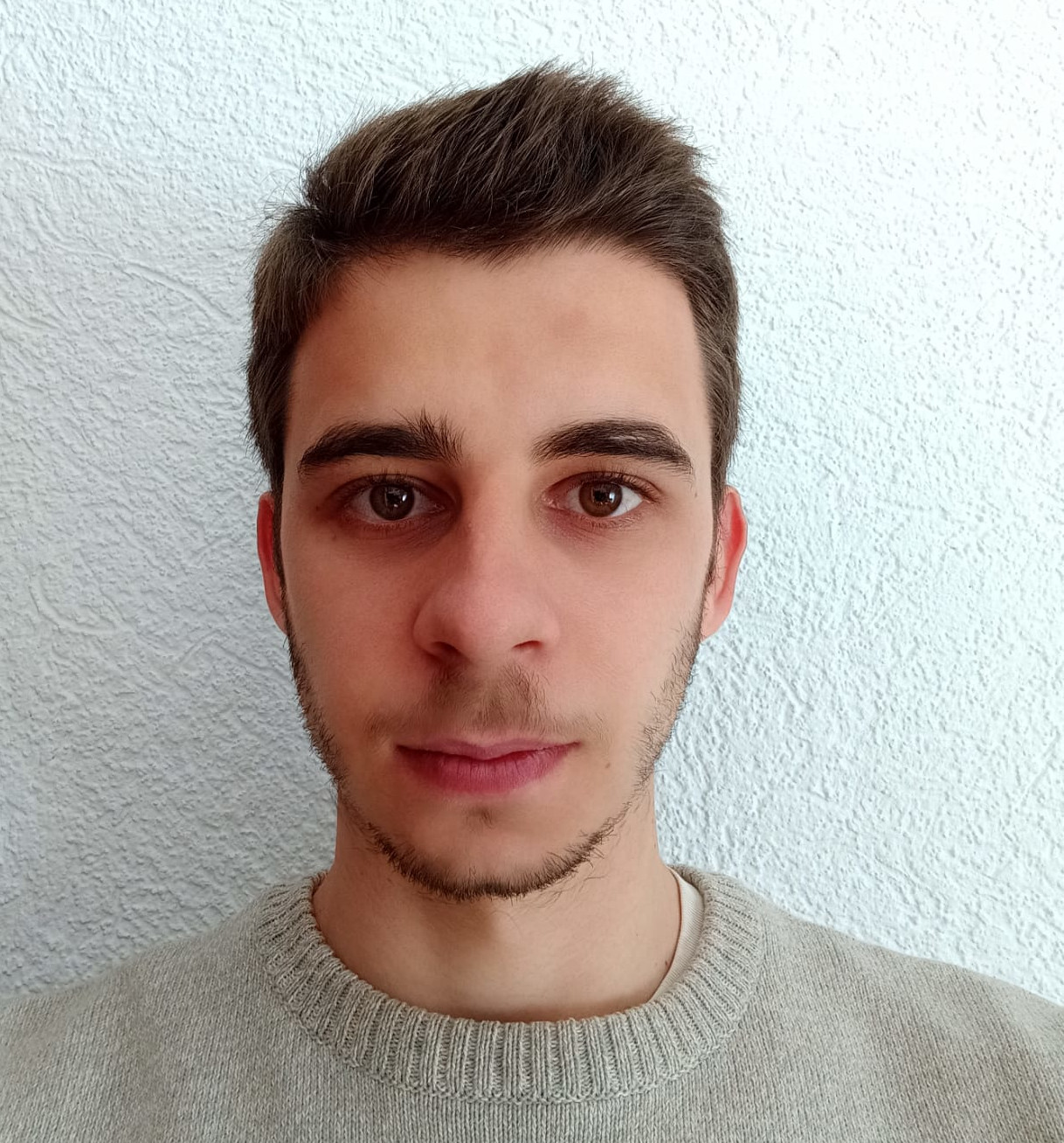}}]{Daniel Coelho} 
received the bachelor’s and M.Sc. degrees from the University of Aveiro, Aveiro, Portugal, in 2019 and 2021, respectively, where he is currently pursuing the Ph.D. degree in mechanical engineering (specialization in artificial intelligence and autonomous driving). His research interests include artificial intelligence, deep learning, reinforcement learning, autonomous driving, localization, and robotics.
\end{IEEEbiography}

\begin{IEEEbiography}[{\includegraphics[width=1in,height=1.25in,clip,keepaspectratio]{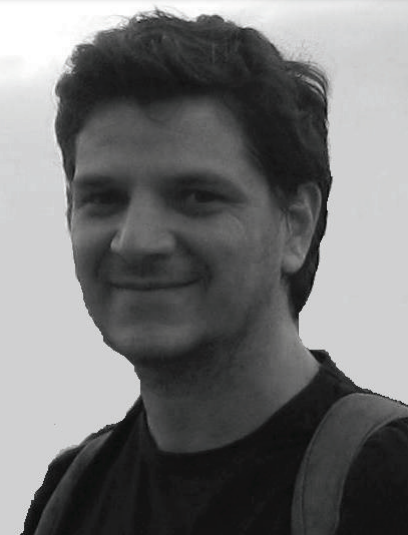}}]{Miguel Oliveira}
received the bachelor’s and M.Sc.
degrees from the University of Aveiro, Aveiro, Portugal, in 2004 and 2007, respectively, and the Ph.D. degree in mechanical engineering (specialization in robotics, on the topic of autonomous driving systems)
in 2013. From 2013 to 2017, he was an Active Researcher with the Institute of Electronics and Telematics Engineering of Aveiro, Aveiro and with the Institute
for Systems and Computer Engineering, Technology and Science, Porto, Portugal, where he participated in several EU-funded projects, as well as national projects. He is currently an
Assistant Professor with the Department of Mechanical Engineering, University of Aveiro, and the director of the Masters in Automation Engineering at that institution. He has supervised more than 20 M.Sc. students. He authored over 20 journal publications from 2015 to 2020. His research interests include autonomous driving, visual object recognition in open-ended domains, multimodal sensor
fusion, computer vision, and the calibration of robotic systems.
\end{IEEEbiography}

\begin{IEEEbiography}[{\includegraphics[width=1in,height=1.25in,clip,keepaspectratio]{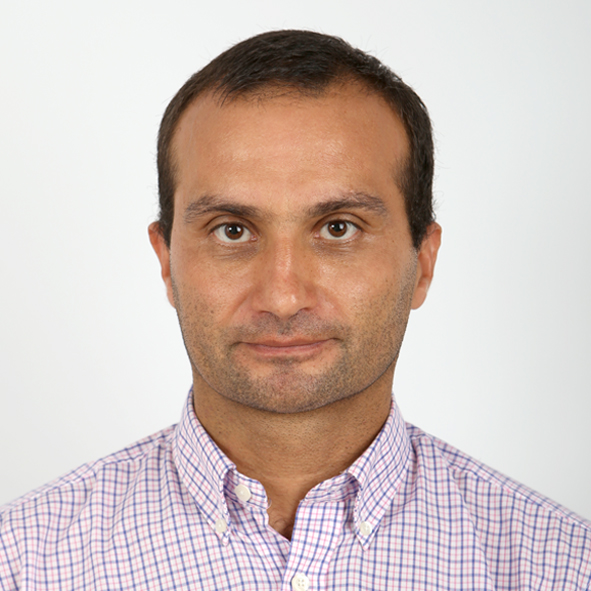}}]{Paulo Dias}
graduated at the University of Aveiro, Portugal in 1998 being Awarded with the Alcatel price “Prémio Engenheiro José Ferreira Pinto Basto” for the best student of Electronics and Telecommunications Engineering of the University of Aveiro. He worked at the Joint research center in Italy from 1998 until 2002. In September 2003, he concluded his PhD at University of Aveiro with the thesis “3D Reconstruction of real-World Scenes Using Laser and Intensity Data”. He is currently an assistant professor within the Department of Electronics Telecommunications and Informatics, University of Aveiro and an active member of IEETA – Institute of Electronics and Telematics Engineering of Aveiro. He is the head of the VARLab, an informal group of individuals within IEETA working in Virtual and Augmented Reality. He has supervised 2 PhD and more than 50 MSc students. He authored over 30 journal publications and 100 Conference papers. His main research interests are 3D Reconstruction, Virtual and Augmented Reality, Computer Vision, Computer Graphics, Visualization, Combination and fusion of data from multiple sensors.
\end{IEEEbiography}

\end{document}